%% file: PaperForReview.tex
\documentclass[10pt,twocolumn,letterpaper]{article}

\usepackage{cvpr}              % To produce the CAMERA-READY version
\usepackage{graphicx}
\usepackage{amsmath}
\usepackage{amssymb}
\usepackage{booktabs}
\usepackage{multirow}
\usepackage{makecell}
\usepackage{tabularx}
\usepackage{bbding}
\usepackage[accsupp]{axessibility} % Improves PDF readability for those with disabilities.
\usepackage[pagebackref,breaklinks,colorlinks]{hyperref}

\usepackage[capitalize]{cleveref}
\crefname{section}{Sec.}{Secs.}
\Crefname{section}{Section}{Sections}
\Crefname{table}{Table}{Tables}
\crefname{table}{Tab.}{Tabs.}

\def\cvprPaperID{4918} % *** Enter the CVPR Paper ID here
\def\confName{CVPR}
\def\confYear{2022}

\begin{document}

\title{Pushing the Performance Limit of Scene Text Recognizer \\ without Human Annotation }

\author{
Caiyuan Zheng$^{1,2}$\thanks{Part of the work was done when C.Zheng was an intern at SRCX.},
Hui Li$^3$, Seon-Min Rhee$^4$, Seungju Han$^4$, Jae-Joon Han$^4$, Peng Wang$^{1,2}$\thanks{P. Wang is the corresponding author.}\\
$^1$School of Computer Science and Ningbo Institute, Northwestern Polytechnical University, China, \\
$^2$National Engineering Laboratory for Integrated Aero-Space-Ground-Ocean \\ Big Data Application Technology, China,\\
$^3$Samsung R\&D Institute China Xi'an (SRCX),\\
$^4$Samsung Advanced Institute of Technology (SAIT), South Korea\\
{\tt\small 2020202704@mail.nwpu.edu.cn, \{hui01.li, s.rhee, sj75.han, jae-joon.han\}@samsung.com}\\
{\tt\small peng.wang@nwpu.edu.cn}
}
\maketitle
%%%%%%%%% ABSTRACT
\begin{abstract}

  Scene text recognition (STR) attracts much attention over the years because of its wide application.
  Most methods train STR model in a fully supervised manner which requires large amounts of labeled data.
  Although synthetic data contributes a lot to STR, it suffers from the real-to-synthetic domain gap that restricts model performance.
  In this work, we aim to boost STR models by leveraging both synthetic data and the numerous real unlabeled images, exempting human annotation cost thoroughly.
  A robust consistency regularization based semi-supervised framework is proposed for STR,
  which can effectively solve the instability issue due to domain inconsistency between synthetic and real images.
  A character-level consistency regularization is designed to mitigate the misalignment between characters in sequence recognition.
  Extensive experiments on standard text recognition benchmarks demonstrate the effectiveness of the proposed method.
  It can steadily improve existing STR models, and boost an STR model to achieve new state-of-the-art results.
  To our best knowledge, this is the first consistency regularization based framework that applies successfully to STR.

\end{abstract}

\input{introduction.tex}
\input{related_work.tex}

\input{method.tex}
\input{experiment.tex}

\section{Conclusion}
In this paper, we propose a robust character-level consistency regularization method for STR.
Our framework consists of a supervised branch trained with synthetic labeled data, and an unsupervised branch trained by two augmented views of real unlabeled images.
An asymmetric structure is designed with EMA, weight decay and domain adaption to encourage a stable model training and overcome the domain gap issue caused by synthetic and real images.
Moreover, a character-level consistency regularization unit is proposed to ensure better character alignment.
Without using any human annotated data, our method is able to improve existing STR models by a large margin, and achieves new SOTA performance on STR benchmarks.

\section*{Acknowledgements}
This work was supported by
National Key R\&D Program of China (No.2020AAA0106900),
the National Natural Science Foundation of China (No.U19B2037, No.61876152),
Shaanxi Provincial Key R\&D Program (No.2021KWZ-03),
Natural Science Basic Research Program of Shaanxi (No.2021JCW-03)
and Ningbo Natural Science Foundation (No.202003N4369).

  %%%%%%%%% REFERENCES
  {\small
    \bibliographystyle{ieee_fullname}
    \bibliography{mybib}
  }

\end{document}

%% file: introduction.tex
%%%%%%%%% BODY TEXT
\section{Introduction}
\label{sec:intro}

\begin{figure}[t]
  \centering
  \begin{subfigure}{0.49\linewidth}
    \includegraphics[width=\textwidth]{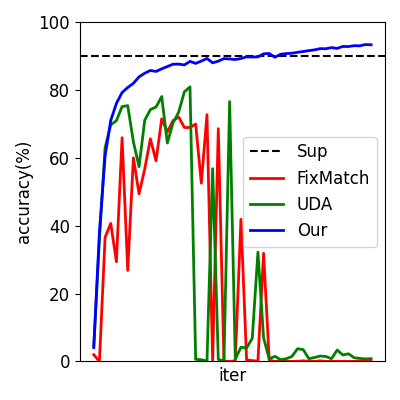}
    \vspace{-6mm}
    \caption{cross-domain.}
    \label{fig:out domain}
  \end{subfigure}
  \hfill
  \begin{subfigure}{0.49\linewidth}
    \includegraphics[width=\textwidth]{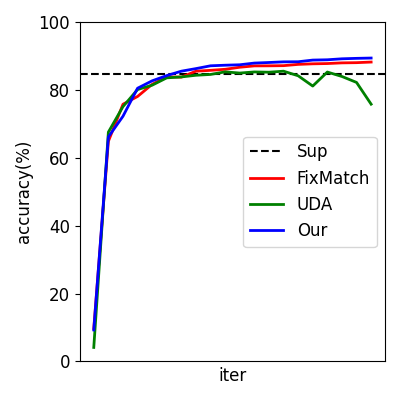}
    \vspace{-6mm}
    \caption{in-domain.}
    \label{fig:in domain}
  \end{subfigure}
  \caption{Scene text recognition test accuracy by using supervised training, existing consistency regularization SSL (UDA~\cite{consitency_regularization/UDA} and FixMatch~\cite{consitency_regularization/FixMatch}) and our method. Cross-domain means the labeled and unlabeled training data are from different domains (\eg. synthetic labeled \vs real unlabeled in our setting), while in-domain means they are from similar condition.
    UDA and FixMatch are feasible in in-domain condition but fail in cross-domain setting.
    It is observed that the test accuracy drops drastically during the training process, and the highest accuracy is even lower than that obtained by supervised training.
    By contrast, our method is able to stabilize the training process and improve test performance in both in-domain and cross-domain conditions. }
  \label{fig:example of in-domain and cross-domain}
  \vspace{-5mm}
\end{figure}

Scene text recognition (STR) is to recognize text in natural scenes and is widely used in many applications
such as image retrieval, robot navigation and instant translation.
Compared to traditional OCR, STR is more challenging because of multiple variations from the environment, various font styles and complicated layouts.

Although STR has made great success, it is mainly researched in a fully supervised manner. Real labeled datasets in STR are usually small because the annotation work is expensive and time-consuming. Hence, two large synthetic
datasets MJSynth~\cite{synthetic_dataset/MJSynth1,synthetic_dataset/MJSynth2} and SynthText~\cite{synthetic_dataset/SynthText} are commonly used to train STR models and produce competitive results.
However, there exists domain gap between synthetic and real data which restricts the effect of synthetic data.
Briefly speaking, synthetic dataset can improve STR performance, but STR model is still hungry for real data.

Considering that it is easy to obtain unlabeled data in real world,
many researchers intend to leverage unlabeled data and train models in a Semi-Supervised Learning (SSL) manner. Baek~\etal~\cite{semisupervised_STR/fewlabels} and Fang~\etal~\cite{semisupervised_STR/ABINet}
introduced self-training methods to train STR models and receive improved performance.
Nevertheless, self-training requires a pre-trained model to predict pseudo-labels for unlabeled data and then re-trains the model, which affects the training efficiency.
By contrast, Consistency Regularization (CR), another important component of state-of-the-art (SOTA) SSL algorithms, has not been well exploited in STR.

In this paper, we would like to explore a CR-based SSL approach to improve STR models, where only synthetic data and unlabeled real data are used for training, exempting human annotation cost thoroughly. CR assumes that the model should output similar predictions when fed perturbed versions of the same image~\cite{consistencyRegul}.
It tends to outperform self-training on several SSL benchmarks~\cite{consistencyRegul2,consistencyRegul3}.
Nevertheless, it is non-trivial to utilize existing CR methods to STR directly.
We attempt to two representative CR approaches, UDA~\cite{consitency_regularization/UDA} and FixMatch~\cite{consitency_regularization/FixMatch}. Neither of them is feasible in our setting. As shown in Figure~\ref{fig:out domain}, the models are quite unstable during the training process.
Compared with experiments on image classification where they show big superiority, we assume the reasons lie in the following two aspects.

1) Our labeled images are synthetic while unlabeled images are from real scenarios.
The domain gap between synthetic and real images affects the training stability. %To be specific, the supervised loss converges normally, but the unsupervised consistency loss is apt to collapse to zero.
Actually, it is found that the collapsed models recognize synthetic inputs with a reasonable accuracy, but generate nearly identical outputs for all real inputs.
We conjecture that they incorrectly utilize the domain gap to minimize the overall loss: they learn to distinguish between synthetic and real data, and learn reasonable representations for synthetic data to minimize the supervised loss, but simply project real data to identical outputs such that the consistency loss is zero.
To validate this conjecture, we perform another experiment by using training images all from real.
As shown in Figure~\ref{fig:in domain}, the training processes of UDA and FixMatch become stable in such a setting.
However, we aim to relieve human labeling cost. The introduced domain gap becomes an issue.

2) Different from image classification, STR is a kind of sequence prediction task.
The alignment between character sequences brings another difficulty to consistency training.

\begin{figure*}[htb]
  \centering
  \includegraphics[width=0.95\linewidth]{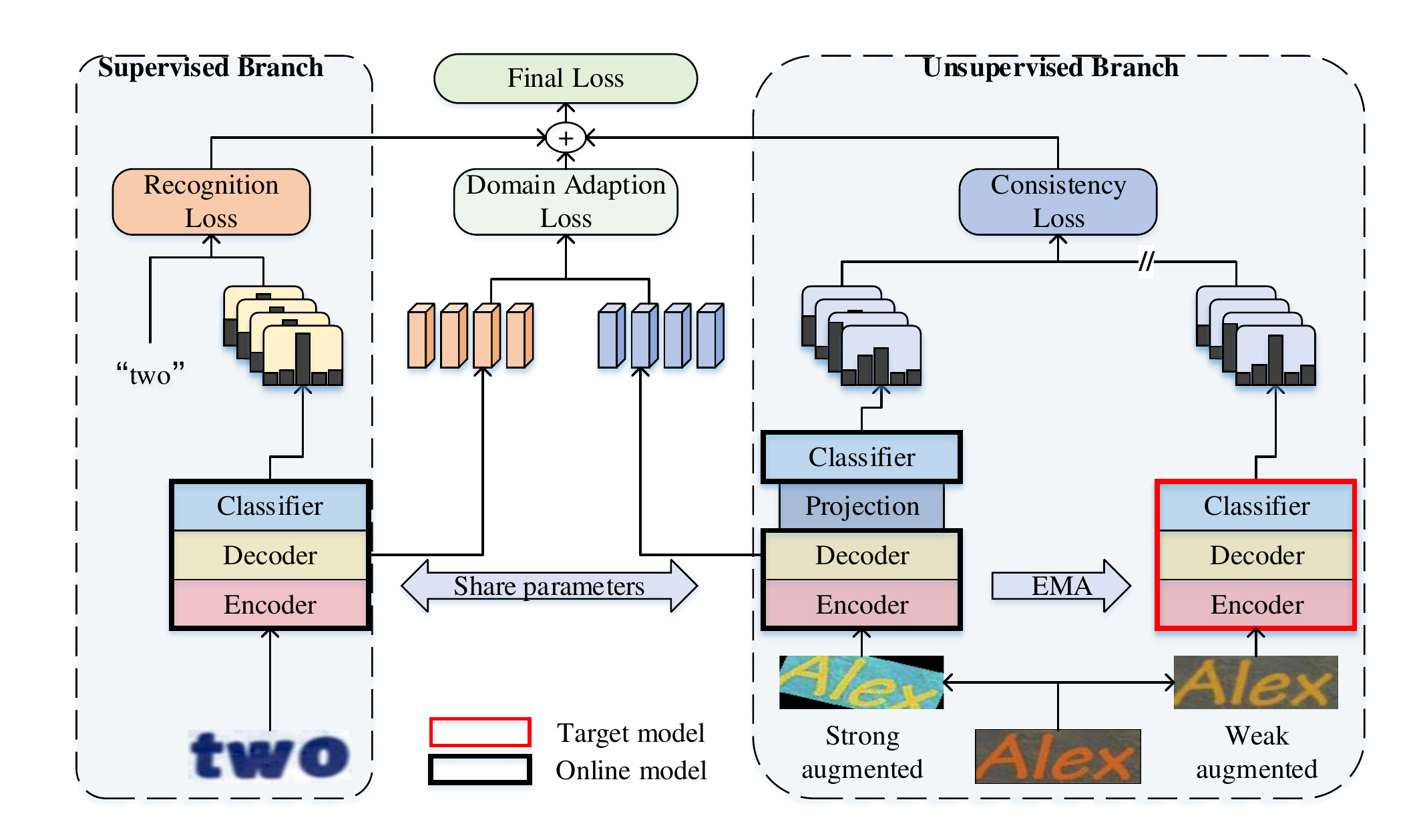}
  \vspace{-2mm}
  \caption{Overall framework of our proposed consistency regularization method for STR. Our model takes advantage of labeled synthetic data and unlabeled real data, exempting human annotation cost thoroughly. An asymmetric structure is designed with EMA and domain adaption to encourage a stable model training. }
  \setlength{\abovecaptionskip}{-0.15cm}
  \label{fig:overall_framework}
  \vspace{-2mm}
\end{figure*}

To address the aforementioned problems, we propose a robust character-level consistency regularization based framework for STR.
Firstly, inspired by BYOL~\cite{self_supervised/BYOL} that prevents model collapse
without using negative samples in contrastive learning, we propose an asymmetric consistency training structure for STR. % to alleviate the instability of the whole model.
Secondly, a character-level CR unit is proposed to ensure the character-level consistency during training process.
Thirdly, some techniques are subtly adopted in training process, such as weight decay and domain adaption, which improve STR model furthermore.

%--------------------------
The main contributions are summarized as follows:

1) We propose a robust consistency regularization based semi-supervised framework for STR.
It is capable of tackling the cross-domain setting, thus more easily benefitting from labeled synthetic data and unlabeled real data.
Compared with self-training approaches, our method is more efficient, without iteratively predicting and re-training.

2) Considering the sequential property of text,
we propose a character-level consistency regularization (CCR) unit to ensure better sequence alignment between the outputs of two siamese models.

3) Extensive experiments are performed to analyze the effectiveness of the proposed framework.
It boosts the performance of a variety of existing STR models.
Despite free of human annotation, our method achieves new SOTA performance on several standard text recognition benchmarks for both regular and irregular text.

%% file: related_work.tex
\section{Related Work}
\label{sec:related work}

%-------------------------------------------------------------------------
\subsection{Scene Text Recognition}
Researches usually treat text recognition as a sequence prediction task and employ RNNs to model the sequences for recognition without character separation. Connectionist temporal classification (CTC) model~\cite{DBLP:conf/nips/wang18,DBLP:journals/pami/ShiBY17} and attention-based encoder-decoder model~\cite{DBLP:conf/cvpr/LeeO16,Rectification/ShiWLYB16} are two commonly used frameworks for STR.
The success of regular
text recognition leads researchers to turn their attention to irregular text recognition.
~\cite{Rectification/LiuCW18,Rectification/ShiWLYB16,Rectification/ShiYWLYB19,Rectification/YangGLHBBYB19,Rectification/ZhanL19,Rectification/LuoJS19}
rectified irregular text into regular ones to alleviate the difficulty in recognition.
~\cite{2Dattention/00130SZ19} and ~\cite{2Dattention/YangWLLZ20}
employed 2D attention to handle the complicated layout of irregular text.
~\cite{attentiondrift/ChengBXZPZ17,attentiondrift/WangZJLCWWC20,attentiondrift/YueKLSZ20}
attempted to improve recognition accuracy by mitigating the alignment drift in attention.
~\cite{semanticinformation/FangXZSTZ18,semanticinformation/QiaoZYZ020,semanticinformation/YuLZLHLD20}
tried to integrate semantic information from language model to enhance word recognition.
All those methods need to be trained in a fully supervised manner.
%-------------------------------------------------------------------------
\subsection{Semi-Supervised Learning}
Semi-Supervised Learning (SSL) aims to use labeled data and additional unlabeled data to boost model performance. There are mainly two types of SSL methods that relate to our work, self-training~\cite{selftraining/entropyminimization,selftraining/noisystudent,selftraining/pseudo-label,selftraining/S4L}
and consistency regularization (CR)~\cite{consitency_regularization/temporal_ensembling,consitency_regularization/meanteacher,consitency_regularization/VAT,consitency_regularization/FixMatch,consitency_regularization/UDA}.
Self-training is simple and effective. It first employs labeled data to train a teacher model, then predicts pseudo labels for unlabeled data, and finally trains a student model using both labeled and pseudo-labeled data. Pseudo Label~\cite{selftraining/pseudo-label}  and Noisy Student~\cite{selftraining/noisystudent} are two popular variants.
CR is based on the manifold assumption that model outputs should be consistent when fed different augmentation views of the same image. For example, Temporal Ensembling~\cite{consitency_regularization/temporal_ensembling} encourages a consensus prediction of
the unknown labels using the outputs of the network-in-training on different epochs.
Mean Teacher~\cite{consitency_regularization/meanteacher}  requires the outputs from teacher model and student model to be consistent, and updates teacher model by averaging student model weights.
FixMatch~\cite{consitency_regularization/FixMatch} combines CR and pseudo-labeling for better performance.
UDA~\cite{consitency_regularization/UDA} argues the importance of noise injection in consistency training, and achieves SOTA performance on a wide variety of language and vision SSL tasks.

\subsection{Semi-Supervised Text Recognition}
Some work has been proposed to train STR model with SSL.
For instance, Gao~\etal~\cite{semisupervised&STR/GaoCWL21} adopted reinforcement learning techniques to exploit unlabeled data for STR performance improvement. However, both labeled and unlabeled data are divided from synthetic data, without domain gap issue.
\cite{semisupervised_STR/scene_domain} and \cite{semisupervised_STR/hand_domain} utilized domain adaption techniques to mitigate the domain shift between source and target data, so as to improve recognition results on target domain.
Baek~\etal~\cite{semisupervised_STR/fewlabels} attempted to train STR model by using real data only, and tried both Pseudo Label and Mean Teacher to enhance STR performance.
Fang~\etal~\cite{semisupervised_STR/ABINet} proposed an autonomous, bidirectional and iterative language modeling for STR. A self-training strategy was applied with the ensemble of iterative prediction to
increase STR performance further.

%% file: method.tex
\section{Proposed Method}
%--------------------------------------------------------
\subsection{Overview}

As shown in Figure~\ref{fig:overall_framework}, our framework consists of an STR model for text recognition and
a CR architecture to integrate information from both labeled and unlabeled data.
We adopt the attention-based encoder-decoder STR model here for illustration. However, our framework is not restricted to autoregressive STR models.
The encoder extracts discriminative features from input images, while the decoder generates character-level features. The classifier maps features into probabilities over character space via a linear transformation and Softmax.

We define two modes for STR model, named training mode and inference mode, according to whether the ``ground-truth'' character sequence is provided.
In training mode, ``ground-truth'' characters are sent to the decoder for next character prediction.
By contrast, in inference mode, the output of previous step is fed into decoder to infer next character. Both modes receive a special ``BOS'' token at the first step which means the start of decoding.
Training mode ends when all ground-truth characters are input, while inference mode ends when generating an ``EOS'' token.

The CR architecture is inspired by UDA~\cite{consitency_regularization/UDA}, which consists of two branches, namely supervised and unsupervised branch, as demonstrated in Figure~\ref{fig:overall_framework}.
The supervised branch is trained on labeled data, while the unsupervised branch takes two augmented views of an unlabeled image as input, and requests the outputs to be similar to each other.
Motivated by BYOL~\cite{self_supervised/BYOL}, we employ STR models with the same architecture but different parameters in unsupervised branch for the two views of inputs, denoted as online model and target model separately.
The online model shares parameters with the one used in supervised branch.
To overcome the instability during model training and improve STR performance,
an additional projection layer is introduced before classifier in online model of the unsupervised branch.

\subsection{Supervised Branch}

Supervised branch adopts the online STR model and runs in training mode, using the labeled synthetic data.
Specially, denote the weight of online STR model as $\theta_{o}$, which is comprised of parameters from three modules, \ie, encoder, decoder and classifier, referring to Figure~\ref{fig:overall_framework}.
Given the input image $\textbf{X}^L$ and the ground-truth character sequence $\textbf{Y}^{gt} = \{ y_1^{gt}, y_2^{gt}, \dots, y_T^{gt} \} $, the supervised branch outputs a sequence of vector $\textbf{P}^{L} = \{ \textbf{p}_1^L, \textbf{p}_2^L, \dots, \textbf{p}_T^L \} $. %A Softmax function is used for each vector $\textbf{z}_t^L$ to transform the output into probability distribution over character space, \ie,
Cross-entropy loss is employed to train the model, \ie,
\vspace{-3mm}
\begin{equation}
  \mathcal{L}_{reg} =  \frac{1}{T} \sum_{t=1}^{T} \log p_t^L(y_{t}^{gt}\mid \textbf{X}^L)
  \vspace{-3mm}
  \label{eq:recognition loss}
\end{equation}
where $p_t^L(y_{t}^{gt})$  represents the predicted probability of the output being $y_{t}^{gt}$ at time step t. T is the sequence length. %and $\theta_{o}$ denotes the parameters of online model used in supervised branch.

\subsection{Unsupervised Branch}
Different from~\cite{consitency_regularization/UDA} and inspired by~\cite{self_supervised/BYOL}, unsupervised branch in our framework relies on two models, referred to as online STR model (with model parameter $\theta_{o}$) and target STR model (with model parameter $\theta_{t}$) respectively. The two models interact and learn from each other.

Given the input image without label $\textbf{X}^U$, two different augmentation approaches are adopted which produce two augmented views of the image, denoted as $\textbf{X}^{U_w}$ and $\textbf{X}^{U_s}$ respectively.
The online STR model takes $\textbf{X}^{U_s}$ as input and runs in training mode.
Motivated by the collapse preventing solution in~\cite{self_supervised/BYOL},
an additional projection layer is introduced between the decoder and classifier, as shown in Figure~\ref{fig:overall_framework}, and the parameters are denoted as $\theta_{p}$ independently. It is composed of 2 layers of perceptron with ReLU activation.
The added projection layer makes the architecture asymmetric between the online and target model, which contributes to a stable training process.
The classifier is then followed to transform the output vector into probabilities over character space, denoted as $\textbf{P}^{U_s} = \{\textbf{p}_1^{U_s}, \textbf{p}_2^{U_s}, \dots, \textbf{p}_T^{U_s} \} $.

The target STR model takes $\textbf{X}^{U_w}$ as input and runs in inference mode, which generates a sequence of probabilities $\textbf{P}^{U_w} = \{\textbf{p}_1^{U_w}, \textbf{p}_2^{U_w}, \dots, \textbf{p}_T^{U_w} \} $. The output sequence is used as the reference target to train the online model.
A stop-gradient operation is acted on the target model, and its parameters $\theta_{t}$ are an exponential moving average (EMA) of the online model parameter $\theta_{o}$, \ie,
\vspace{-1mm}
\begin{equation}
  \theta_t = \alpha\theta_t + (1-\alpha)\theta_o
  \vspace{-1mm}
  \label{eq:EMA update}
\end{equation}
where $\alpha \in [0,1]$ is the target decay rate.
EMA makes the target model produce relatively stable targets for online model, which helps to keep the projection layer in near optimal and benefits the model training as well.

As indicated in~\cite{selftraining/entropyminimization,consitency_regularization/UDA},
regularizing predictions with low entropy would be beneficial to SSL. We sharpen the output from target STR model $\textbf{P}^{U_w}$ by using a low Softmax temperature $\tau$.
Denote the output vector at step $t$ before Softmax as $\textbf{z}_t^{U_w} = \{z_1^{U_w}, z_2^{U_w},\dots, z_C^{U_w} \}$, $C$ is the number of character classes, then
\vspace{-2mm}
\begin{equation}
  p_t^{U_w}(y_t)=\frac {\exp(z^{U_w}_{y_t}/\tau)} {{\sum_{y^\prime_t} \exp(z^{U_w}_{y^\prime_t}/\tau)}}
  \vspace{-2mm}
  \label{eq:caculate confident score}
\end{equation}
We set $\tau=0.4$ following~\cite{consitency_regularization/UDA}.

\begin{figure}[t!]
  \centering
  \includegraphics[width=1.0\linewidth]{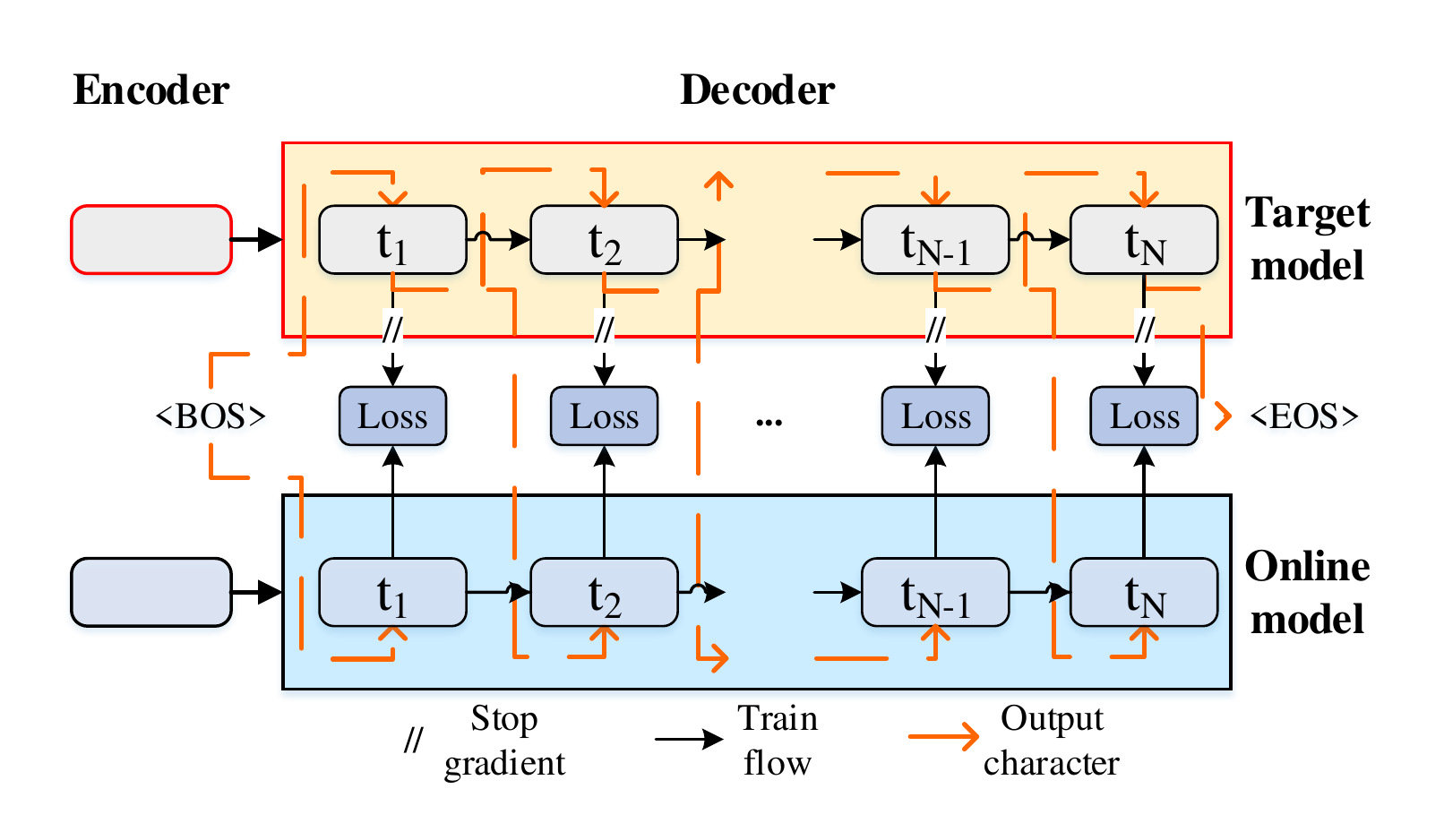}
  \vspace{-6mm}
  \caption{Character-level consistency regularization (CCR).
    In each time step t, target decoder and online decoder share the same output character produced by target decoder in previous time step so as to keep good character alignment.
    Consistency loss is computed between the outputs in each time step.}
  \setlength{\abovecaptionskip}{-0.15cm}
  \label{fig:character-level consistency regularization}
  \vspace{-1mm}
\end{figure}

The consistency training regularizes the outputs of $\textbf{P}^{U_w}$ and $\textbf{P}^{U_s}$ to be invariant. However, given that STR is a sequence recognition task, a character-level consistency regularization (CCR) unit is proposed for autoregressive decoder, so as to keep a good sequence alignment.
As shown in Figure~\ref{fig:character-level consistency regularization}, %instead of running separately, the online STR model will takes the pseudo label generated by target model
in decoding time step $t$, a pseudo label is generated from target model by taking the class that has the highest probability in $\textbf{p}_t^{U_w}$. The pseudo label will be used as the input for both online and target decoder in next time step.
The design enforces online decoder and target decoder share the same context information, benefits character level alignment, and thus ensures a stable consistency training.

To alleviate the influence caused by noise samples in training process, we filter out noise samples based on their confidence scores in recognition.
The confidence score is the cumulative product of the maximum output probability from target model in each decoding step, \ie,
\vspace{-3mm}
\begin{equation}
  \mathcal{\textbf{s}}^{U_w} = \prod_{t=1}^{T}p_t^{U_w}(y_t \mid \textbf{X}^{U_w})
  \vspace{-3mm}
  \label{eq:confscore}
\end{equation}
The consistency loss used in unsupervised branch is then defined as:
\vspace{-3mm}
\begin{equation}
  \mathcal{L}_{cons} =  \mathbb I (\mathcal{\textbf{s}}^{U_w}>\beta_{U}) \frac{1}{T} \sum_{t=1}^{T} Dist(\textbf{p}_t^{U_w}, \textbf{p}_t^{U_s})
  \vspace{-3mm}
  \label{eq:cons}
\end{equation}
where $ \mathbb I (\mathcal{\textbf{s}}^{U_w}>\beta_{U})$ is an indicator, $\beta_{U}$ is a threshold for filtering out noises and
$Dist(\cdot)$ is a function to measure the character-level distance between $\textbf{P}^{U_w}$ and $\textbf{P}^{U_s}$.
There are several choices for $Dist$, such as Cross Entropy (CE), KL-divergence or Mean Squared Error (MSE).
KL-divergence is adopted in our framework by default.

\subsection{Additional Training Techniques}
\textbf{Weight Decay.}
Weight decay is an important component in contrastive learning~\cite{simclr,self_supervised/BYOL} and SSL~\cite{consitency_regularization/FixMatch}.
It is claimed that~\cite{weight_decay} weight decay in BYOL can help balance weights between predictor and online model dynamically, and improve the representation ability of online model. Here we also adopt it into our model training so as to improve the feature learning capability of online model. %, which will be used in inference time.

\textbf{Domain Adaption.}
To mitigate the domain shift in training data, a character-level domain adaptation unit is employed between the supervised and unsupervised branches, referring to~\cite{semisupervised_STR/scene_domain}.
Specially, in each decoding step, decoder of the online model extracts vision feature for the character to be decoded,
denoted as $\textbf{H}^L = \{\textbf{h}^L_1, \textbf{h}^L_2, \cdots, \textbf{h}^L_T\}$
and $\textbf{H}^{U_s} = \{\textbf{h}^{U_s}_1, \textbf{h}^{U_s}_2, \cdots, \textbf{h}^{U_s}_T\}$
for features extracted in supervised and unsupervised branch respectively.
Domain adaption loss is defined as

\vspace{-2mm}
\begin{equation}
  \mathcal{L}_{da} = \frac{1}{4d^2}\Vert(cov(\textbf{H}^L)-cov(\textbf{H}^{U_s})\Vert^2_F
  \vspace{-2mm}
  \label{eq: domain adaption loss}
\end{equation}
where
$\Vert\cdot\Vert^2_F$ denotes the squared matrix Frobenius norm, $cov(\textbf{H})$ is covariance matrix,
$d$ is the feature dimension.

\subsection{Overall Objective Function}

We sum the three loss functions defined above. The overall objective function for training our proposed model is:
\begin{equation}
  \mathcal{L}_{overall} = \mathcal{L}_{reg} + \lambda_{cons}\mathcal{L}_{cons}+\lambda_{da}\mathcal{L}_{da}
  \label{eq:overall loss}
\end{equation}
where $\lambda_{cons}$ and $\lambda_{da}$ are hyper-parameters to balance three terms.
We set $\lambda_{cons}=1$ and $\lambda_{da}=0.01$ empirically.

%% file: experiment.tex
\section{Experiment}

\subsection{Datasets}

Two types of data are used here for training, \ie, synthetic data with annotations and real data without label.

Two widely used synthetic datasets are adopted including \textbf{SynthText (ST)}~\cite{synthetic_dataset/SynthText} and \textbf{MJSynth (MJ)}~\cite{synthetic_dataset/MJSynth2}, which results in $14.5$M samples in total, referring to as \textbf{synthetic labeled data (SL)}.

For real unlabeled scene text data, we collected from three public available datasets, Places2~\cite{Places2}, OpenImages\footnote{https://storage.googleapis.com/openimages/web/index.html} and ImageNet ILSVRC 2012~\cite{ImageNet}. CRAFT~\cite{CRAFT} was employed to detect text from these images.
Then we cropped text images with detection scores larger than $0.7$.
Images with low resolution (width times height is less than $1000$) were also discarded.
There are finally $10.5$M images, denoted as \textbf{real unlabeled data (RU)}.

In addition, during ablation study, to demonstrate the superiority of the proposed framework, we also conduct experiments by using real labeled data collected by~\cite{semisupervised_STR/fewlabels}. It has $278$K images totally, named as \textbf{real labeled data (RL)}.

%\subsubsection{Test Data}
Six commonly used scene text recognition benchmarks are adopted to evaluate our method. %, including both regular and irregular ones.

\textbf{ICDAR 2013 (IC13)} contains $1095$ cropped word images. Following ~\cite{semanticinformation/YuLZLHLD20}, we remove images that contain non-alphanumeric characters, which results in 857 test patches.

\textbf{IIIT5K-Words (IIIT)}~\cite{IIIT5k} has $3000$ nearly horizontal word patches for test.

\textbf{Street View Text (SVT)}~\cite{DBLP:conf/iccv/WangBB11} consists of $647$ word images collected from Google Street View for test.

\textbf{SVT-Perspective (SVTP)}\cite{SVTP)} contains $645$ images for test, which are cropped from side-view snapshots in Google Street View.

\textbf{CUTE80 (CUTE)}~\cite{CUTE} has $288$ curved text images.

\textbf{ICDAR 2015 (IC15)}~\cite{IC15} contains $2077$ word images cropped from incidental scene images. After removing images with non-alphanumeric characters, there are $1811$ word patches left for test.

\subsection{Evaluation Metric}

Following common practice, we report word-level accuracy for each dataset. Moreover, in order to comprehensively evaluate models for their recognition performance on both regular and irregular text, following~\cite{semisupervised_STR/fewlabels}, we introduce an average score (Avg) which is the accuracy over the union of samples in all six datasets.

\subsection{Implementation Details}

The whole model is trained end-to-end without pre-training.
We use a batch size of $384$ for labeled data and $288$ for unlabeled data.
By default, we set the target decay rate $\alpha=0.999$ and
confidence threshold $\beta_{U}=0.5$ respectively.
Both supervised branch and unsupervised branch are jointly trained, while we only use the model in supervised branch in inference time.

Four STR models are adopted to validate the effectiveness of the proposed framework, with their default model configurations,
including CRNN~\cite{DBLP:journals/pami/ShiBY17}, MORAN~\cite{MORAN}, HGA~\cite{2Dattention/YangWLLZ20} and TRBA~\cite{TRBA}. Note that CRNN uses CTC for character decoding, which is non-autoregressive. Hence, CCR is not adopted when training model with CRNN.

We adopt Adadelta when training MORAN or HGA, following their original optimization method. The learning rate is $1.0$ initially and decreases during training process.
AdamW~\cite{AdamW} optimizer  is adopted when using CRNN or TRBA model.
Following~\cite{semisupervised_STR/fewlabels}, we use the one-cycle learning rate scheduler~\cite{Super-convergence} with a maximum
learning rate of $0.001$. The weight decay rate is aligned with the used STR model.

The unsupervised branch takes two augmented views of an image as input. Here we define two types of augmentations, \ie, StrongAug and WeakAug.
StrongAug is borrowed from RandAugment~\cite{Randaugment} which includes multiple augmentation strategies on both geometry transformations and color jitter.
Considering Cutout may crop some characters from the image which will corrupt the semantic information of text, we remove "Cutout" operation from RandAugment.
WeakAug only has color jitter, including brightness, contrast, saturation and hue.
In our framework, we use WeakAug for target model and StrongAug for online models of both supervised and unsupervised branches.

\begin{table*}[ht!]
  \newcommand{\tabincell}[2]{\begin{tabular}{@{}#1@{}}#2\end{tabular}}
  \begin{center}
    \scalebox{0.8}{
      \begin{tabular}{ccccccccccc}
        \toprule
        \multirow{2}{*}{} & Methods                                                  & \multirowcell{2}{Labeled                                                                                                                                                   \\Dataset} & \multirowcell{2}{Unlabeled\\Dataset} & \multicolumn{3}{c}{Regular Text} & \multicolumn{3}{c}{Irregular Text} & \multirow{2}{*}{Avg}                                                                 \\

                          &                                                          &                          &               & IC13             & SVT              & IIIT             & IC15             & SVTP          & CUTE             &                  \\
        \midrule
        \multirow{14}{*}{\rotatebox{90}{SOTA Methods}}
                          & Shi~\etal~\cite{DBLP:journals/pami/ShiBY17} (CRNN)       & MJ                       & -             & -                & 80.8             & 78.2             & -                & -             & -                & -                \\
                          & Luo~\etal~\cite{MORAN}(MORAN)                            & SL                       & -             & -                & 88.3             & 93.4             & 77.8             & 79.7          & 81.9             & -                \\
                          & Yang~\etal~\cite{2Dattention/YangWLLZ20}(HGA)            & SL                       & -             & -                & 88.9             & 94.7             & 79.5             & 80.9          & 85.4             & -                \\
                          & Baek~\etal~\cite{TRBA}(TRBA)                             & SL                       & -             & -                & 87.5             & 87.9             & -                & 79.2          & 74.0             & -                \\
                          & Liao~\etal~\cite{MaskTextSpotter}(Mask TextSpotter)      & SL                       & -             & 95.3             & 91.8             & 93.9             & 77.3             & 82.2          & 87.8             & 88.3             \\
                          & Wan~\etal~\cite{TextScanner}(TextScanner)                & SL                       & -             & 92.9             & 90.1             & 93.9             & 79.4             & 84.3          & 83.3             & 88.5             \\
                          & Wang~\etal~\cite{attentiondrift/WangZJLCWWC20}(DAN)      & SL                       & -             & 93.9             & 89.2             & 94.3             & 74.5             & 80.0          & 84.4             & 87.2             \\
                          & Yue~\etal~\cite{attentiondrift/YueKLSZ20}(RobustScanner) & SL                       & -             & 94.8             & 88.1             & 95.3             & 77.1             & 79.5          & \underline{90.3} & 88.4             \\
                          & Qiao~\etal~\cite{semanticinformation/QiaoZYZ020}(SRN)    & SL                       & -             & 95.5             & 91.5             & 94.8             & 82.7             & 85.1          & 87.8             & 90.4             \\
                          & Zhang~\etal~\cite{SPIN}(SPIN)                            & SL                       & -             & -                & 90.9             & 95.2             & 82.8             & 84.3          & 83.2             & -                \\
                          & Mou~\etal~\cite{PlugNet}(PlugNet)                        & SL                       & -             & -                & 92.3             & 94.4             & -                & 84.3          & 84.3             & -                \\
                          & Qiao~\etal~\cite{PIMNet}(PIMNet)                         & SL                       & -             & 95.2             & 91.2             & 95.2             & 83.5             & 84.3          & 84.4             & 90.5             \\
                          & Fang~\etal~\cite{semisupervised_STR/ABINet}(ABINet)      & SL                       & -             & \underline{97.4} & \underline{93.5} & \underline{96.2} & \underline{86.0} & 89.3          & 89.2             & \underline{92.7} \\
        \cline{2-11}
                          & Gao~\etal~\cite{DBLP:journals/tip/GaoCWL21}              & $10\%$  SL               & $90\%$  SL    & -                & 78.1             & 74.8             & -                & -             & -                & -                \\
                          & Baek~\etal~\cite{semisupervised_STR/fewlabels}(CRNN)     & RL                       & Book32 et al. & -                & 84.3             & 89.8             & -                & 74.6          & 82.3             & -                \\
                          & Baek~\etal~\cite{semisupervised_STR/fewlabels}(TRBA)     & RL                       & Book32 et al. & -                & 91.3             & 94.8             & -                & 82.7          & 88.1             & -                \\
                          & Fang~\etal~\cite{semisupervised_STR/ABINet}(ABINet)      & SL                       & Uber-Text     & 97.3             & 94.9             & \textbf{96.8}    & 87.4             & 90.1          & 93.4             & 93.5             \\

        \midrule
        \midrule
        \multirow{9}{*}{\rotatebox{90}{Ours}}
                          & CRNN-pr                                                  & SL                       & -             & 91.0             & 82.2             & 90.2             & 71.6             & 70.7          & 81.3             & 82.8             \\
                          & CRNN-cr                                                  & SL                       & RU            & 92.4             & 87.9             & 92.0             & 75.8             & 75.7          & 85.8             & 85.9             \\
        \cline{2-11}
                          & MORAN-pr                                                 & SL                       & -             & 95.1             & 90.4             & 93.4             & 79.7             & 80.6          & 85.4             & 88.5             \\
                          & MORAN-cr                                                 & SL                       & RU            & 96.5             & 93.0             & 94.1             & 82.6             & 82.9          & 88.5             & 90.2             \\
        \cline{2-11}
                          & HGA-pr                                                   & SL                       & -             & 95.0             & 89.5             & 93.6             & 79.8             & 81.1          & 87.8             & 88.7             \\
                          & HGA-cr                                                   & SL                       & RU            & 95.4             & 93.2             & 94.9             & 84.0             & 86.8          & 92.0             & 91.2             \\
        \cline{2-11}
                          & TRBA-pr                                                  & SL                       & -             & 97.3             & 91.2             & 95.3             & 84.2             & 86.4          & 92.0             & 91.5             \\
                          & TRBA-cr                                                  & $10\%$ SL                & $10\%$ RU     & 97.3             & 94.7             & 96.2             & 87.0             & 89.6          & \textbf{94.4}    & 93.2             \\
                          & TRBA-cr                                                  & SL                       & RU            & \textbf{98.3}    & \textbf{96.3}    & 96.5             & \textbf{89.3}    & \textbf{93.3} & 93.4             & \textbf{94.5}    \\
        \bottomrule
      \end{tabular}
    }
    \caption{Comparison with SOTA methods on STR test accuracy.
      In each column, the best result is shown in bold, and the best result in supervised setting is
      shown with underline. "-pr" means our reproduced results and "-cr" means using our \textbf{c}onsistency \textbf{r}egularization method. Our method improves STR models firmly, and propels TRBA towards new SOTA performance on test benchmarks.}
    \label{comparison with SOTA}
  \end{center}
  \vspace{-6mm}
\end{table*}

\subsection{Comparison with SOTA}

We perform experiments by using different STR models.
For fair comparison, we also reproduce those models under supervised setting using the same data augmentation strategy as that used in our semi-supervised training.
As presented in Table~\ref{comparison with SOTA}, our reproduced models have comparable or even higher accuracies than that reported in the original paper. Those results provide an even fair baseline to show the advantage of our method. Experiments with their original settings can be found in Supplementary.

By training with the proposed framework using additional unlabeled real images, all models gain improvement.
To be specific, CRNN improves by $3.1\%$ (from $82.8\%$ to $85.9\%$) on average, MORAN increases from $88.5\%$ to $90.2\%$ (+1.7\%).
HGA has an accuracy increase of $2.5\%$ (from $88.7\%$ to $91.2\%$) and TRBA has an increase of $3.0\%$ (from $91.5\%$ to $94.5\%$).
The consistent enhancement over different STR models shows the effectiveness and universality of our proposed method.
Specially, the performance gain over irregular text (IC15, SVTP and CUTE) is more obvious, since irregular text has more variance on appearance which is hard to be generated by synthetic engine.

Note that although TRBA is worse than ABINet~\cite{semisupervised_STR/ABINet}  in supervised setting ($91.5\%$ \vs $92.7\%$), our framework helps TRBA outperform ABINet that adopts self-training in semi-supervised setting ($94.5\%$ \vs $93.5\%$), which proves the superiority of our proposed CR method again. Compared with other SOTA work, our proposed framework with TRBA achieves the highest accuracies on vast majority of test datasets (only except IIIT), which demonstrates its robustness for both regular and irregular text recognition.

In addition, to accelerate training process, we perform an experiment with TRBA using only $10\%$ synthetic labeled data (denoted as ``$\text{SL}_{sm}$'' that contains only $1.45$M images) and $10\%$ real unlabeled data (denoted as ``$\text{RU}_{sm}$'' which has $1.05$M images). Surprisingly, experimental results is fairly good with the average score of $93.2\%$, even higher than that obtained by $\text{TRBA}_{pr}$ ($91.5\%$) and ABINet~\cite{semisupervised_STR/ABINet} ($92.7\%$). It should be noted that $\text{TRBA}_{pr}$ and ABINet are trained in a fully supervised manner using all synthetic data ($14.5$M). The training data is $5.8$ times more than that used in $\text{TRBA}_{sm}$. The excellent results suggest the necessary of using real images in training STR models and the advantage of our semi-supervised training framework.

\begin{figure}[t!]
  \centering
  \includegraphics[width=1.0\linewidth]{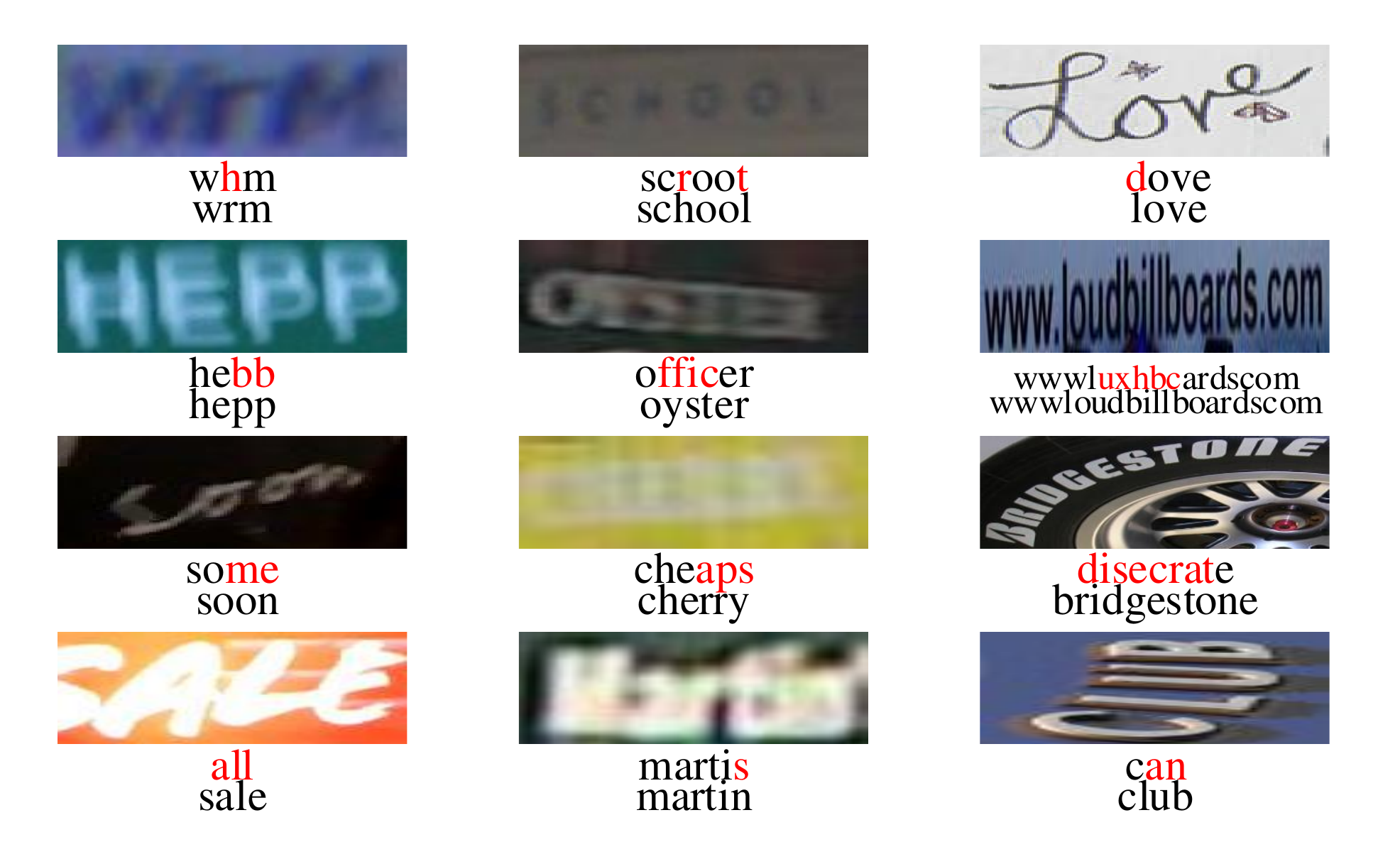}
  \vspace{-5mm}
  \caption{Hard examples that can be successfully recognized by using our method. The first line shows the recognition results by $\text{TRBA}_{pr}$, which include mistakes (red characters), while the second line are results by  $\text{TRBA}_{cr}$. Our method enables TRBA to address even tough samples like dark, blur, or severe distortion.
  }
  \label{fig:vis}
  \vspace{-3mm}
\end{figure}

In Figure~\ref{fig:vis}, we present several examples that can be correctly recognized by $\text{TRBA}_{cr}$ but encounter failure when using $\text{TRBA}_{pr}$.
Although the employed real images are unlabeled, STR models can still get benefit from our method, particularly for recognizing text that is severely blurred, distorted, or with artistic font.

\subsection{Ablation Study}

In order to analyze the proposed model,
we conduct a series of ablation experiments in this section. All ablation experiments are performed using TRBA because of its good performance.  $\text{SL}_{sm}$ and $\text{RU}_{sm}$ are employed for fast training. More experiments with different data sizes can be found in Supplementary.
\vspace{-4mm}
\subsubsection{Effect of domain gap on model stability}
\vspace{-1mm}
In this work, we propose a stable CR based SSL framework for STR.
As stated in Section~\ref{sec:intro}, we guess it is the domain inconsistency among training data used in STR that causes the instability or even failure by previous CR methods.

To prove this conjecture, we perform experiments using domain consistent training data (in-domain data).
Specially, we split the real labeled training data RL into $\text{RL}_{20p}$ and $\text{RL}_{80p}$
with a ratio of 1:4. $\text{RL}_{20p}$ is adopted with labels while $\text{RL}_{80p}$ is employed without annotations.
SOTA CR methods are tested, including FixMatch~\cite{consitency_regularization/FixMatch} and UDA~\cite{consitency_regularization/UDA}.
As presented in Table~\ref{tab: in domain and cross domain}, when training data is from the same domain, they work well.
The test accuracy increases by $3.6\%$ using FixMatch and $2.6\%$ using UDA.
However, when the training data is from different domains, \eg, $\text{SL}_{sm}$ and $\text{RU}_{sm}$, their training processes become unstable. We test the models before collapse. The recognition accuracies are even lower than that obtained by only using $\text{SL}_{sm}$, with performance degradation of  $11.0\%$ (FixMatch) and $4.6\%$ (UDA) separately. % compared to supervised training result.

By contrast, our method is able to improve the recognition accuracy no matter the training data is from similar domain or not.
In comparison to the results by fully supervised training, our method improves STR model accuracy steadily by $4.5\%$ ($84.8\%$ to $89.3\%$) using in-domain data and $3.3\%$ ($89.9\%$ to $93.2\%$) in cross-domain setting.
The performance gain in in-domain setting is even larger than that brought by FixMatch and UDA.

\begin{table}
  \newcommand{\tabincell}[2]{\begin{tabular}{@{}#1@{}}#2\end{tabular}}
  \begin{center}
    \setlength{\tabcolsep}{1.8mm}
    \scalebox{0.75}{
      \begin{tabular}{ccccccc}
        \toprule
                                                      & \tabincell{c}{Labeled/                                                        \\ Unlabeled Data}       & Methods & \tabincell{c}{IC13 \\IC15} & \tabincell{c}{SVT\\ SVTP} &\tabincell{c}{IIIT\\CUTE}& Avg  \\
        \midrule
        \multirow{4}{*}{\rotatebox{90}{In-domain}}    & \tabincell{c}{$\text{RL}_{\text{20p}}$(55.7K)/                                \\- }                &Sup                         & \tabincell{c}{90.1 \\77.6 }     & \tabincell{c}{87.5 \\78.0 } & \tabincell{c}{88.8 \\83.0 } &84.8 \\
        \cline{2-7}                                   & \multirowcell{3}{$\text{RL}_{\text{20p}}$(55.7K)/                             \\$\text{RL}_{\text{80p}}$(223K) }    & FixMatch        & \tabincell{c}{93.0 \\82.3 }     & \tabincell{c}{88.6 \\82.5 } & \tabincell{c}{92.0 \\88.5 } &88.4 \\
        \cline{3-7}                                   &                                                   & UDA  & \tabincell{c}{92.5 \\80.7 }     & \tabincell{c}{88.6 \\80.9 } & \tabincell{c}{91.4 \\88.5 } &87.4 \\
        \cline{3-7}                                   &                                                   & Ours & \tabincell{c}{93.8 \\82.5 }     & \tabincell{c}{91.5 \\83.6 } & \tabincell{c}{92.9 \\88.5 } &89.3 \\
          \hline\hline
        \multirow{4}{*}{\rotatebox{90}{Cross-domain}} & \tabincell{c}{$\text{SL}_{\text{sm}}$(1.45M)/                                 \\-                  } &Sup                        & \tabincell{c}{96.0 \\82.4 }     & \tabincell{c}{90.0 \\82.6 } & \tabincell{c}{94.4 \\88.9 } &89.9 \\
        \cline{2-7}                                   & \multirowcell{3}{$\text{SL}_{\text{sm}}$(1.45M)/                              \\$\text{RU}_{\text{sm}}$(1.06M) }   &FixMatch         & \tabincell{c}{90.0 \\72.6 }     & \tabincell{c}{86.2 \\77.2 } & \tabincell{c}{79.2 \\69.1 } &78.9 \\
        \cline{3-7}                                   &                                                   & UDA  & \tabincell{c}{94.2 \\75.7 }     & \tabincell{c}{85.3 \\79.5 } & \tabincell{c}{90.0 \\82.3 } &85.3 \\
        \cline{3-7}                                   &                                                   & Ours & \tabincell{c}{97.3 \\87.0 }     & \tabincell{c}{94.7 \\89.6 } & \tabincell{c}{96.2 \\94.4 } &\textbf{93.2} \\
        \bottomrule
      \end{tabular}
    }
    \caption{Experiments with CR methods on in-domain and cross-domain data settings. Our method can consistently improve recognition accuracy. The results of FixMatch and UDA in cross-domain setting are obtained by the models before collapse. }
    \label{tab: in domain and cross domain}
  \end{center}
  \vspace{-10mm}
\end{table}

\vspace{-3mm}
\subsubsection{Ablation on model units}
\vspace{-1mm}

The techniques used in our method
include an additional projection layer for asymmetric structure, EMA, domain adaption and weight decay.
Here we analyze the effect of each unit in detail. The experiments are performed with CCR added to benefit character-level consistency.

As presented  in Table~\ref{tab:ablation_for_PCT}, the use of additional projection layer can improve the final average score by $0.7\%$.
However, the performance is still lower than that obtained under fully supervised setting ($87.7\%$ \vs $89.9\%$).
As indicated in~\cite{weight_decay}, without weight decay, the consistency between online and target outputs is dependent mainly on the projection layer, rendering the online model weights inferior.
Weight decay helps balance weights between online model and projection layer dynamically.
The use of weight decay, with projection layer, increases the average score on test data by another $3.5\%$, surpassing the supervised results.
EMA mechanism brings an accuracy gain of $1.6\%$ furthermore as it helps keep projection layer in near-optimal and improves training stability.
Lastly, the adding of domain adaption improves the average test accuracy up to $93.2\%$.

\begin{table}
  \newcommand{\tabincell}[2]{\begin{tabular}{@{}#1@{}}#2\end{tabular}}
  \begin{center}
    \scalebox{0.8}{
      \begin{tabular}{cccccccc}
        \toprule
        Projection     & WD             & EMA            & DA             & \tabincell{c}{IC13 \\IC15} & \tabincell{c}{SVT\\ SVTP} &\tabincell{c}{IIIT\\CUTE}& Avg  \\
        \midrule
                       &                &                &                & \tabincell{c}{94.2 \\80.5 } & \tabincell{c}{91.5\\84.0 } & \tabincell{c}{88.7 \\84.0 } &87.0\\
        \cline{5-8}
        \CheckmarkBold &                &                &                & \tabincell{c}{94.5 \\81.6 } & \tabincell{c}{90.1\\86.1 } & \tabincell{c}{89.5 \\85.4 } &87.7\\
        \cline{5-8}
        \CheckmarkBold & \CheckmarkBold &                &                & \tabincell{c}{97.2 \\85.9 } & \tabincell{c}{93.0\\87.0 } & \tabincell{c}{93.5 \\91.3 } &91.2\\
        \cline{5-8}
        \CheckmarkBold & \CheckmarkBold & \CheckmarkBold &                & \tabincell{c}{96.7 \\86.7 } & \tabincell{c}{94.6\\89.3 } & \tabincell{c}{95.9 \\92.7 } &92.8\\
        \cline{5-8}
        \CheckmarkBold & \CheckmarkBold & \CheckmarkBold & \CheckmarkBold & \tabincell{c}{97.3 \\87.0 } & \tabincell{c}{94.7\\89.6 } & \tabincell{c}{96.2 \\94.4 } &\textbf{93.2}\\
        \bottomrule
      \end{tabular}
    }
    \caption{Ablation on model units.
      ``Projection'' means using additional projection layer before classifier. "WD" means weight decay,
      "EMA" means using EMA for target model. "DA" means domain adaption.}
    \label{tab:ablation_for_PCT}
  \end{center}
  \vspace{-5mm}
\end{table}

\begin{table}
  \newcommand{\tabincell}[2]{\begin{tabular}{@{}#1@{}}#2\end{tabular}}
  \begin{center}
    \setlength{\tabcolsep}{3mm}
    \scalebox{0.8}
    {
      \begin{tabular}{ccccc}
        \toprule
        Method & \tabincell{c}{IC13 \\IC15} & \tabincell{c}{SVT\\ SVTP} &\tabincell{c}{IIIT\\CUTE}& Avg  \\
        \midrule
        SCR    & \tabincell{c}{96.6 \\84.9 } & \tabincell{c}{93.0 \\85.9 } & \tabincell{c}{96.4 \\93.1 } &92.2\\
        \hline
        CCR    & \tabincell{c}{97.3 \\87.0 } & \tabincell{c}{94.7 \\89.6 } & \tabincell{c}{96.2 \\94.4 } &\textbf{93.2}\\
        \bottomrule
      \end{tabular}
    }
    \caption{Effect of our proposed CCR. Compared to using standard consistency regularization, training with CCR conduces to $1\%$ average score increase for TRBA.}
    \label{tab:comparison CCR with SCR}
  \end{center}
  \vspace{-5mm}
\end{table}

\begin{table}
  \newcommand{\tabincell}[2]{\begin{tabular}{@{}#1@{}}#2\end{tabular}}
  \begin{center}
    \setlength{\tabcolsep}{3mm}
    \scalebox{0.8}{
      \begin{tabular}{cccccccc}
        \toprule
        Consistency Loss & \tabincell{c}{IC13 \\IC15} & \tabincell{c}{SVT\\ SVTP} &\tabincell{c}{IIIT\\CUTE}& Avg  \\
        \midrule
        MSE              & \tabincell{c}{96.3 \\84.0 } & \tabincell{c}{92.0 \\86.8 } & \tabincell{c}{94.2 \\92.0 } &91.0\\
        \hline
        CE               & \tabincell{c}{97.4 \\86.9 } & \tabincell{c}{94.3 \\89.8 } & \tabincell{c}{96.3 \\92.7 } &\textbf{93.2}\\
        \hline
        KL-divergence    & \tabincell{c}{97.3 \\87.0 } & \tabincell{c}{94.7 \\89.6 } & \tabincell{c}{96.2 \\94.4 } &\textbf{93.2}\\
        \bottomrule
      \end{tabular}
    }
    \caption{Ablation on different distance functions used in consistency loss. CE and KL-divergence leads to similar performance, better than MSE.}
    \label{tab:consistency_loss}
  \end{center}
  \vspace{-5mm}
\end{table}

\vspace{-5mm}
\subsubsection{Effect of CCR}
\vspace{-2mm}
Another contribution of this work is a character-level consistency regularization (CCR) unit to handle the specially sequential property of STR task.
Instead of letting online model and target model run separately in unsupervised branch (standard consistency regularization, SCR), and only restricting their final outputs by consistency loss,
we proposed CCR to enforce the same context information for both online and target model.
Experimental results in Table~\ref{tab:comparison CCR with SCR} prove the effectiveness of CCR.
It helps TRBA improve $1\%$ more on the final test accuracy.

\subsubsection{Ablation on distance measure functions}
By default, we use KL-divergence to measure the consistency in loss function (\ref{eq:cons}).
Here we test other distance measure functions, such as CE and MSE. As presented in Table~\ref{tab:consistency_loss}, empirically, CE leads to similar recognition performance with KL-divergence, while MSE results in lower accuracies ($93.2\%$ \vs $91.0\%$).

\subsection{Comparison with Other Semi-supervised Methods}

We compare our method with other SSL approaches that have been successfully used in STR, including Pseudo Label (PL)~\cite{selftraining/pseudo-label} and Noisy Student (NS)~\cite{selftraining/noisystudent}. TRBA is used as the basic model. %, and $\text{SL}_{sm}$, $\text{RU}_{sm}$ are employed as training data.
PL based SSL is performed following the practice in~\cite{semisupervised_STR/fewlabels},
while NS based SSL is following~\cite{selftraining/noisystudent}, with the threshold $\beta_U=0.5$ and $3$ iterations of re-training.

The results are shown in Table~\ref{tab:semi supervised comparison}. Our CR based method outperforms all the others, with the resulted average score $2.3\%$ higher than PL and $0.8\%$ higher than NS. Note that compared to NS, our training process is more efficient, without time-consuming iterations.

\begin{table}
  \newcommand{\tabincell}[2]{\begin{tabular}{@{}#1@{}}#2\end{tabular}}
  \begin{center}
    \setlength{\tabcolsep}{3mm}
    \scalebox{0.8}{
      \begin{tabular}{ccccccc}
        \toprule
        Method             & \tabincell{c}{IC13 \\IC15} & \tabincell{c}{SVT\\ SVTP} &\tabincell{c}{IIIT\\CUTE}& Avg  \\
        \midrule
        Pseudo Label (PL)  & \tabincell{c}{95.9 \\82.9 } & \tabincell{c}{91.2 \\85.7 } & \tabincell{c}{95.4 \\90.6 } &90.9\\\hline
        Noisy Student (NS) & \tabincell{c}{96.3 \\85.5 } & \tabincell{c}{94.4 \\86.7 } & \tabincell{c}{96.1 \\94.1 } &92.4\\\hline
        Ours               & \tabincell{c}{97.3 \\87.0 } & \tabincell{c}{94.7 \\89.6 } & \tabincell{c}{96.2\\94.4 } &\textbf{93.2}\\
        \bottomrule
      \end{tabular}
    }
    \caption{Comparison with other semi-supervised methods. Our method brings more benefit to STR model and outperforms the other approaches.}
    \label{tab:semi supervised comparison}
  \end{center}
  \vspace{-8mm}
\end{table}
% -----------------------------------------------------------------------